\documentclass[runningheads]{llncs}
\usepackage{esvect} %to display vector
\usepackage[usenames, dvipsnames]{color} 
\usepackage[linesnumbered,ruled]{algorithm2e} % for algorithm
\usepackage{url}

\usepackage{graphicx}
\usepackage{epstopdf}
\usepackage[caption=false]{subfig}
\begin{document}
\title{Improving short text classification through global augmentation methods}
\author{
Vukosi Marivate\inst{1,2}\orcidID{0000-0002-6731-6267} \and
Tshephisho Sefara\inst{2}\orcidID{0000-0002-5197-7802}
% \authorrunning{
% F. Author et al.
\institute{
University of Pretoria, South Africa \and
Council for Scientific and Industrial Research, South Africa \\
\email{vukosi.marivate@cs.up.ac.za}\\
\email{tsefara@csir.co.za}
}
}
\maketitle              % typeset the header of the contribution
\begin{abstract}

We study the effect of different approaches to text augmentation. To do this we use 3 datasets that include social media and formal text in the form of news articles. Our goal is to provide insights for practitioners and researchers on making choices for augmentation for classification use cases. We observe that Word2vec-based augmentation is a viable option when one does not have access to a formal synonym model (like WordNet-based augmentation). The use of \emph{mixup} further improves performance of all text based augmentations and reduces the effects of overfitting on a tested deep learning model. Round-trip translation with a translation service proves to be harder to use due to cost and as such is less accessible for both normal and low resource use-cases.

\end{abstract}
\section{Introduction}
In this paper, we look at data augmentation for Natural Language Processing (NLP) applications. Encouraged by the ever-present use of data augmentation in computer vision applications \cite{cubuk2018autoaugment}, we want to be able to provide researchers and practitioners with a better understanding of augementation for NLP tasks. Augmentation has benefited many image classification tasks \cite{cubuk2018autoaugment}, the structure of images are changed in order to increase the number of samples available to the machine learning algorithm while introducing resilience in the final model. Augmentation of text data to be able to create robust models has had different factors of success. Some approaches require more direct information about the language at hand than others~\cite{li2017robust},  while others are more agnostic given a learned language model that can be used~\cite{kobayashi2018contextual}.

Over the last few years, the development of distributed word representations (word embeddings)~\cite{mikolov2013distributed} has improved the modelling of semantic relations between words in a text. This has created many new approaches to understanding text, in the case of this work, text classification tasks. In order to improve the classification accuracy, as well as make models more robust, one can look at data augmentation as a way to improve performance and create robustness. More recently, the development of unsupervised language models \cite{peters2018deep,howard2018universal}, makes it possible for us to use more data-driven language models that can be combined with data augmentation methods to improve performance and robustness of machine learning models for NLP. 

We are motivated by a number of factors. We would like to be able to train classification models that do not necessarily have a large amount of labelled data. Labelling of data comes at a cost. Whether it is for identifying fake news \cite{krishnan2018identifying}, understanding political phenomena \cite{ratkiewicz2011detecting}, feedback on government services \cite{sano2015automatic}, or better coordination during emergencies \cite{imran2015processing}, getting labels is always challenging. As such, knowing that building machine learning models requires a large amount of data, we need to still be able to build models with smaller data. A large organisation might have access to large data sets as well as resources to label a large chunk of it. A smaller organisation tends to not have large data and fewer resources to label. To extend the use of the classifiers further than the distribution of information fed into it, we need to be able to change the input data in a way that makes the final learned model more robust to slight changes in the input distribution. This could be caused by the evolution of language or even geographical changes. Another use is in semi-supervised learning, where we use the few labels we have to create a classifier (that is likely noisy) to label more unlabelled data and then feed this back to train another classifier.  

Our contributions, in this paper, is a short survey of a number of data augmentation methods, specifically looking at methods that augment data with more of a global view. That is, the schemes replace words that are used similarly from a global view instead of a contextually local view. So how are similar words used across texts, instead of what might be the best word to replace in this specific sentence in this specific document. We discuss methods that use linguistic features, a model that uses a translation service and then augmentation methods that act on embeddings/language models. To better understand the behaviour of the augmentation methods, we evaluate the approaches on a number of classification datasets under a number of conditions and provide insights into the different approaches. We also show the effect of the \emph{mixup} \cite{zhang2017mixup} method on NLP tasks as an augmentation approach.  This paper is organised as follows; We cover different text augmentation methods first. The approach in our comparative study is described in Section~\ref{sec:approach}. Section~\ref{sec:results} discusses the experimental results and then we conclude in Section~\ref{sec:conclusion}.

\section{Augmentation methods for text\label{sec:background}}
% There are different data augmentation techniques useful for controlling generalisation error for machine learning models, namely, synonym-based, translation-based, and word embeddings/language models. 

\begin{figure}[ht!]
\centering
 \includegraphics[width=0.95\textwidth]{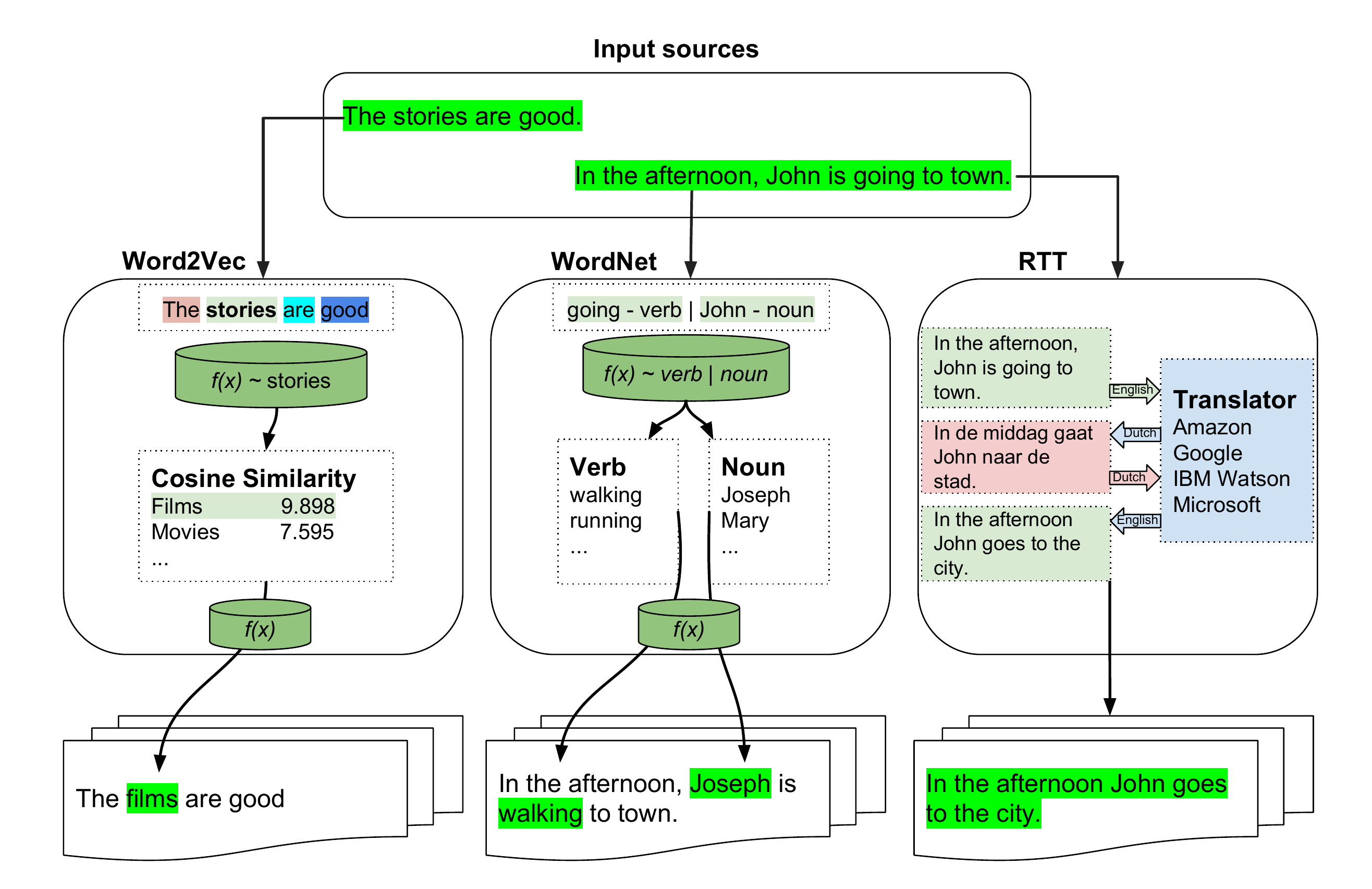}
    \caption{Overview of augmentation structures\label{fig:augment_diagram}}
\end{figure}

For many machine learning tasks, data augmentation has been employed as a regularisation method while training supervised machine learning models. The more diverse examples fed to the model during training, the better the model generalises, and consequently, the better they predict when presented with new examples. Data augmentation is famously used in images, audio and more recently in text. In this section, we describe prior approaches to text augmentation. For our work, we categorise text augmentation techniques into two categories. Namely, \textit{augmentation on text source} and \textit{augmentation on text representation}. In the rest of this section, we discuss methods that fall in either of these categories.  A summary of the methods we discuss in this section is shown in Table~\ref{tab:summary_table} and Fig.~\ref{fig:augment_diagram} illustrate how these methods augment data. 

\begin{table*}[ht!]
\centering
\caption{Augmentation techniques\label{tab:summary_table}}
 \begin{tabular}{|l|l|l|l|l|} 
 \hline
 Method & Augmentation on  & Labels & Linguistics & Semantic \\ %[0.5ex] \hline
 \hline
 Synonyms from WordNet \cite{zhang2015character} & \textit{text source} & No & Yes & Yes \\ 
 Words from Word2vec \cite{kobayashi2018contextual} & \textit{text source} & No & Yes & Yes \\ 
%  Noise Addition \cite{xie2017data} & \textit{?} & No & ? & ? \\
 Round-trip Translation \cite{lauclarklappin2015,sennrich2016improving} & \textit{text source} & No & Yes & Yes \\
 \emph{mixup} \cite{zhang2017mixup} & \textit{text representation} & Yes & No & No \\
%  Contextual Augmentation \cite{kobayashi2018contextual} & ? & No & ?& ? \\ [1ex] 
 \hline
 \end{tabular}
\end{table*}

\subsection{Augmentation on source text} Augmenting textual data requires language modelling rules, written by linguistic experts, which are limited for low-resource languages. The alternative way is to use a dictionary/thesaurus or database of synonyms to make word replacements. In a situation where there is no dictionary, the other way is to use distributed word representation where semantically similar words are identified together. In this subsection, we present how word replacements and sentence paraphrasing can be used as augmentation techniques.   
% Newish paper Contextual Augmentation: Data Augmentation by Words with Paradigmatic Relations
\subsubsection{Synonym Augmentation}
We first begin with methods that use linguistic features. The best way to augment data is by using human rephrasing of sentences, but this is expensive. Therefore, the most natural option in data augmentation for most authors is to replace words or phrases with their synonyms. Verbs and nouns are the best name classes to have synonyms in various context. The popular open-source lexical database for the English language is WordNet \cite{miller1995wordnet}. It groups words like nouns, verbs, adjectives and adverbs into sets of cognitive synonyms called \textit{synsets} each expressing different concept, provides short definitions and usage examples, and records a number of relations among these synonym sets. WordNet thus superficially resembles a thesaurus in that it groups words together based on their meanings. However, with the distinction that WordNet labels the semantic relations among words and it interlinks word forms and strings of letters specific to senses of words resulting with words found in close proximity to one another in the network semantically disambiguated. A good illustration of synonym augmentation is shown in Fig.~\ref{fig:augment_diagram} as WordNet. 

The thesaurus-based augmentation method has been applied to training a Siamese adaptation of the Long Short-Term Memory (LSTM) network to assess the semantic similarity between sentences \cite{mueller2016siamese}. Zhang \textit{et al}. \cite{zhang2015character} used a WordNet-based augmentation method to augment their training data by using a Geometric function to help select words from a given dataset, using the selected words to find their synonyms to train a temporal convolutional network that learns text understanding from character level input up to abstract text concepts. 

\subsubsection{Semantic Similarity Augmentation}

Using distributed word representation (word embeddings)  \cite{mikolov2013distributed}, one can identify semantically similar words~\cite{kobayashi2018contextual}. This approach requires either pre-trained word embedding models for the language at hand or enough data from the target application to be able to build the embedding model. This approach thus does not require access to a dictionary or thesaurus for a language in order to find synonyms. This can benefit languages where such resources might be harder to obtain but there might be enough unsupervised text data to be able to build the embedding models. With recent advances in building complete language models \cite{radford2018improving,devlin2018bert,peters2018deep,fedus2018maskgan}, further advances can be made to identify syntactic and semantic similarity for word augmentation. 

For this paper, we are not exploring the use of language models. One can view language models as enabling a more localised replacement of words given the local context in the sentence. On the other hand, word embeddings give a global context. Language Models such as that by \cite{devlin2018bert} allow the filling in of blanks in any part of a sentence. While other language models can be more used to fill in the \emph{next} missing word in a sentence\cite{radford2018improving}, reading a sentence left to right then filling in the missing word at the end. This exploration remains an area of future work.

\subsubsection{Round-trip Translation}
Round-trip translation (RTT) is also known as recursive, back-and-forth, and bi-directional translation. It is the process of translating a word, phrase or text into another language (forward translation), then translating the results back into the original language (back translation) \cite{aiken2010efficacy}. RTT can be used as an augmentation technique to increase the training data. An example is shown in Fig.~\ref{fig:augment_diagram}. 
Fadaee \textit{et al}. \cite{fadaee2017data} used a method called translation data augmentation that uses rare word substitution to augment data for low-resourced neural machine translation (NMT). Their method augments both the source and target sentence to create a new pair preserving the meaning of the sentence. They proposed a weaker notion of label preservation that allows alteration of sentence pairs as long as the sentences remain translations of each other. 

Sennrich \textit{et al}. \cite{sennrich2016improving} used back translation as a data augmentation method to leverage the target monolingual data. In their method, both the NMT model and training algorithm are kept unaltered, and they employed back translation to construct training data. That is, target monolingual sentences are translated with an existing machine translation system into source language, that is used as additional parallel data to retrain the source-to-target NMT model. Aroye hun and Gelbukh \cite{aroyehun2018aggression} used English RTT on four intermediate languages (French, Spanish, German, and Hindi) that increased the size of their training set by a factor of 5. This method improved the performance of their model for identification of aggression in social media posts. Although back-translation has been proven to be effective and robust \cite{aroyehun2018aggression}, one major problem for further improvement is the quality of automatically generated training data from monolingual sentences. Some of the incorrect translations are very likely to decrease the performance of a source-to-target model due to the deficiency of a machine translation system.

\subsection{Augmentation on representation}

Augmentation on source text requires that we are able to have access to a method to replace words in the source text to create more examples while at the same time keeping the meaning of the full sentence the same. In this subsection, we present a method that was introduced as a regularisation technique on Deep Neural Networks but can be viewed as augmentation that instead acts on the text representation. We discuss \emph{mixup} which acts both on the input and the outputs.

\subsubsection{\textit{mixup} Augmentation}
 
\textit{mixup}, introduced in \cite{zhang2017mixup} can be seen as an augmentation method that can be classified as \textit{augmentation on representation}.  \emph{mixup} is data agnostic and is applied to images, speech and tabular data in the original paper \cite{zhang2017mixup}.  \emph{mixup} creates new training examples by drawing samples (sets of two or more) from the original data and combining them convexly. It combines the data both in terms of the input and output. The simplest implementation, which the authors suggest, creates a new augmented dataset by taking pairs of samples from the initial dataset and convexly summing both the input and output. That is, given two examples $(X_1,y_1), (X_2,y_2)$, where $X$ is the input vector and $y$ is the one-hot encoded labels, we construct a new example:

$$\hat{X} = \lambda X_1 + (1 - \lambda) X_2$$
$$\hat{y} = \lambda y_1 + (1 - \lambda) y_2$$ 

\noindent where $\lambda \in [\ 0,1 ]\ $ and is drawn from Beta distribution, $\lambda = \beta(\alpha,\alpha)$, $\alpha$ is set \cite{zhang2017mixup}. It is somewhat akin to soft-labelling but is best suited for learning algorithms that use the cross-entropy loss and also changes the input. The standard method is used for classification problems (it remains further work to explore the uses for regression). For text problems, augmentation by \textit{mixup} can be done on the text representation. As such we can use \textit{mixup} with bag-of-words models, TFIDF \cite{ramos2003using}, word embeddings and language models. 

\section{Method and approach}\label{sec:approach}
Our goal is to measure the impact of augmentation methods on the performance of classification algorithms. We would like to compare methods across different datasets and on as level a playing field as possible. It is important to do this so that insights can be used by other researchers as they build tools for different classification problems in text. We first describe the data we will use for classification, then the learning algorithms we will be using, and finally the last subsection details the experiments we will perform to test the augmentation approaches to better understand their strengths and limitations.

\subsection{Data}

The data we will use is from three different datasets that represent different challenges with text. For short text, we use the Sentiment 140 data set \cite{go2009twitter} as well as the Hate Speech data set \cite{hateoffensive}. For longer text, we use the AG News data set used in \cite{zhang2015character}. The Sentiment 140 uses noisy labelling~\cite{go2009twitter} in the form of weak supervision  to label the training data. The Hate Speech dataset uses human annotated labels. Even though the Sentiment 140 has a separate test set that is made up of human annotated labels, we will first focus on splitting the original data. The data is summarised in Table \ref{table:data}. 
\begin{table}[ht]
\centering
\caption{Summary of data\label{table:data}}
    \begin{tabular}{|l|l|l|l|l|}
    \hline
    Data Source & Words & Approx. Sentences & Avg Number of Words & Labels\\%[0.1ex]\hline 
    \hline
    Sentiment 140 & 18M & 2M & 13 & Binary\\
    AG News & 3M & 154K & 31 & Categorical\\
   Hate Speech & 243K & 24K & 14 & Categorical\\[1ex]
    \hline 
    \end{tabular}
\end{table}

\subsection{Augmentation algorithms}
We augment each data set using four types of augmentation methods: WordNet-based augmentation, Word2vec-based augmentation, Round-trip translation as well as \textit{mixup}. The first three methods augment data on text and can be combined with \textit{mixup} which augments data on the fly. As part of this paper, we also release a library that allows the use of all text-based augmentation methods\footnote{\url{https://github.com/dsfsi/textaugment}}

\subsubsection{WordNet-based synonym augmentation}
WordNet-based augmentation is a type of augmentation that randomly selects $r$ words from a sentence (if they exist) using any distribution. To decide on which words to replace, we selected replaceable words like verbs, nouns, and the combination of them using a part-of-speech tagger implemented in the natural language toolkit (NLTK)\cite{bird2009natural} from a given text and randomly selects $r$ of them. The probability of $r$ is calculated by a Geometric distribution $P[r]\sim p^r$ where parameter $p$ is the probability of success. The synonym $s$ chosen given a word is also determined by another Geometric distribution in which $P[s]\sim p^s$ using the same probability of success $p=0.5  $\cite{zhang2015character}. The new sentence is constructed by replacing the selected verb or noun with their synonyms. The algorithm has options to choose to augment using either verbs or nouns or even a combination. We report the results choosing the outcome of $p$ on the first trial. 
\subsubsection{Word2vec-based (learned semantic similarity) augmentation}
Word2vec is another robust augmentation method that uses a word embedding model \cite{mikolov2013distributed} trained on the public dataset to find the most similar words for a given input word. We use both a pretrained Wikipedia Word2Vec model for formal text. For social media data, we convert a Glove model, pretrained on Twitter data, to Word2vec format using Gensim \cite{rehurekGensimlrec}. We load the converted models in our algorithm to augment data by randomly selecting a word in a sentence to determine its similar words using cosine similarity. To select a similar word, we use the cosine similarity as a relative weight to select a similar word that replaces the input word. Our algorithm is illustrated in Algorithm \ref{alg:word2vec}, it receives a string and an integer where the string is an input data and the integer represents a number of repetitions to augment a given input data. The advantage of Word2vec is that it tends to produce vectors that are more topically related, in other words it allows words with similar meaning to have similar representation. 
\begin{algorithm}
\SetAlgoLined
\SetKwFunction{Aug}{Augment}
\SetKwProg{Fn}{def}{\string:}{}
\SetKwFunction{Range}{range}%%
\SetKw{KwTo}{in}\SetKwFor{For}{for}{\string:}{}%
\SetKwIF{If}{ElseIf}{Else}{if}{:}{elif}{else:}{}%
\SetKwFor{While}{while}{:}{fintq}%
\KwIn{$s$: a sentence, $run$: a number}
\KwOut{$\hat{s}$ a sentence with words replaced}
\Fn{\Aug{s,run}}{
 Let $\vv{V}$ be a vocabulary\;
%  Let R $\gets Random(s)$ be a random distribution\;
 \For{ $i$ \KwTo \Range{run} }{
    % $w_i \gets R(s) $ \tcc*[r]{ select word}
    $w_i \gets$ randomly select a word from $s$\;
    % $ s_0 \gets \arg \max \limits_{\substack{v\in \vv{V}}} cosine(w_i,v)$\tcc*[r]{ get word}
    $\vv{w} \gets$ find similar words of $w_i$\;
    $s_0 \gets$ randomly select a word from $\vv{w}$ given weights as distance\;
    $\hat{s} \gets $replace $w_i$ with similar word $s_0$\;
%  $\hat{s} \gets s_0$\tcc*[r]{ replace word}
 }
 return($\hat{s}$)\;
}
\caption{Word2vec-based augmentation algorithm \label{alg:word2vec}\cite{wang2015s}.}
\end{algorithm}
\subsubsection{Round Trip Translation (RTT) augmentation}
We implement RTT augmentation using Google translation services\footnote{\url{http://translate.google.com}} as well as Amazon translate\footnote{\url{https://aws.amazon.com/translate/}}. We translate text in English to a target language then back to English. To measure the effect of RTT on a different number of augmentations, we translated text to 
% Spanish, 
French
% , Latin, Dutch, and Afrikaans
then back to English. We ensured that the paraphrased back-translated texts carry the same meaning as the source text. Due to the cost of doing the augmentation, we are unable to show results on our larger datasets. This is a challenge in using RTT.
% Table~\ref{tab:rtt} provides an example of RTT method.
% \begin{table}[hb]
%     \centering
%     \caption{Examples of augmented data using RTT\label{tab:rtt}}
%     \begin{tabular}{|l| p{6cm}|}
%         \hline
%          Language &Sentence  \\
%          \hline\hline
%          English:&quality distribution is hammered after reporting a large loss for the second quarter\\[0.1cm]
%          Spanish:&la distribución de la calidad se golpea después de informar una gran pérdida para el segundo trimestre\\[0.1cm]
%          \textbf{English}:&\textbf{the distribution of quality hits after reporting a large loss for the second quarter}\\[0.1cm]
%          \hline
%     \end{tabular}
% \end{table}
\subsubsection{\textit{Mixup} augmentation} We implement this method with the $\alpha$ variable set to 0.2. We run \textit{mixup} on its own as well as in combination with the other methods described.

\subsubsection{Evaluation metrics}
For all of the experiments, we use error, and with some experiments loss, as metrics to give insight to the behaviour of the augmentation algorithms and their impact on model performance. We use the $error$ for the rest of the paper defined as.
\begin{equation}
    error = 1 - accuracy 
\end{equation}

We also present cross-entropy loss when investigating overfitting.

\subsection{Learning Algorithms}

We test two types of learning algorithms for the classification of text. Specifically, we use a deep learning approach as well as a logistic regression (LR). The inputs of the algorithms will be slightly different given the different types of augmentation. We would like to be able to show the impact of augmentation on the state of the art approaches as well as the impact on traditional algorithms and NLP representation. 

%deine the models and maybe their equations
We implement our models using Keras\footnote{\url{http://keras.io}} (a deep learning toolkit). For logistic regression, the model contains a dense layer given few arguments; a number of labels, and activation function as \textit{softmax}.
% , and both L1 and L2 kernel regularizers set to  $0$ and $0.1$ respectively.
We compile the model using an adaptive learning rate method for gradient descent called ADADELTA \cite{zeiler2012adadelta} as our optimiser, and categorical cross entropy as our loss function. For the DNN, we use a Bidirectional LSTM network \cite{hochreiter1997long} coupled with 1 dimensional convolutional neural network  and 3 fully connected layers activated by a \textit{rectified linear unit} and \textit{softmax} illustrated in Table.~\ref{table:dnn}.

\begin{table}[ht]
\centering
\caption{Table of a DNN architecture. Dropout layers use probability of 0.4. Tanh represents a rectified linear unit. The number of filters for the last layer corresponds to the number of classes of the given dataset. \label{table:dnn}}
    \begin{tabular}{|l|l|l|l|l|}
    \hline
    Layer & Type & Filters/Neurons & Kernel \\%[0.1ex] \hline 
    \hline 
    % 0 & Embedding & ? & ?& ?\\
    1 & Bidirectional LSTM+tanh & 256 & - \\
    2 & Dropout & - & - \\
    3 & Conv1D+sigmoid & 512 & 3 \\
    4 & GlobalMaxPool1D & 512 & - \\
    5 & Fully connected+tanh & 512 & - \\
    6 & Dropout & - & - \\
    7 & Fully connected+tanh & 200 & - \\
    8 & Dropout & - & - \\
    9 & Fully connected+softmax & 4 & - \\
    \hline 
    \end{tabular}
\end{table}

% \subsection{System Specifications}
% We use Amazon AWS to conduct the experiments on DNN, the environment is configured with one NVIDIA Tesla K80 GPU shown in Table~\ref{table:specs}. For LR, we configure our environment with two Tesla K20 GPUs and two GeForce GTX690 GPUs on Ubuntu desktop with 12 Intel i7 CPUs at 3.5 GHz. 
% \begin{table}[ht]
% \centering
% \caption{System specification on Amazon AWS\label{table:specs}}
%     \begin{tabular}{|l|l|l|l|l|}
%     \hline
%     Hardware & Specification \\%[0.1ex] \hline 
%     \hline 
%     System & Ubuntu x86\_64\\
%     GPU Memory & 11441MiB \\
%     CPU & 4 $\times$ Intel(R) Xeon(R) CPU E5-2686 v4 @ 2.30GHz \\ 
%     GPU & NVIDIA Tesla K80 with CUDA v10 \\
%     \hline 
%     \end{tabular}
% \end{table}
\section{Experiments and Results\label{sec:results}}

This section explains detailed experiments on the effectiveness of augmentation in several settings. We conduct three types of experiments on the datasets. The first experiment tests the importance of augmentation on limited data. The second experiment tests the effectiveness of augmentation under overfitting settings. The last experiment tests how the number of augmentations affects performance. 
% The last experiment tests the effectiveness of augmentation on noisy data. 

\subsection{Effect of augmentation on less data}
Limited data is a challenge both in getting well-labelled data for supervised learning problems \cite{dundar2015simplicity} and also continues to be a bigger challenge for low-resource languages \cite{das2017named}.
%as well as in situations where. 
Here we present what the effect of the different augmentation schemes are when labelled data is reduced. 

Here we show the effect of augmentation on learning for logistic regression and Deep Neural Network (DNN) models. For logistic regression, we use a TFIDF vector scheme \cite{ramos2003using}. For TFIDF, we fit the vector model using only the training data. For the DNN, we use pre-trained word vector representations from Glove \cite{pennington2014glove}. For Glove we use the pre-trained Wikipedia model for the AG News dataset and pre-trained Twitter model for the social media datasets. For all of the datasets in this experiment, we augment the data 5 times. We also include the original dataset, resulting in an augmented dataset that is 6 times the size of the original. To make results comparable, when we use no augmentation we just repeat the same original dataset 6 times. Due to limitations of RTT we were only able to run experiments on logistic regression on AG News and on the Hate Speech Dataset. As we cannot compare RTT across all experiments, we discuss it at the end.
% For the experiment we focus on using the logistic regression algorithm as the learning algorithm. 
% % Input is used as ...... 
% We compare both using TFIDF vector scheme  and using Glove . For the TFIDF scheme, we test the performance of the models using TFIDF only trained on the available training data, not on the full dataset. For the Glove implementation, we use pre-trained word vectors. 
% % (as such also look at the transfer learning performance).

\begin{figure}[ht!]
\centering
    \subfloat[AG News - LR error\label{fig:augmentation_results_lr_error_ag_news}]{%
        \includegraphics[width=0.45\textwidth]{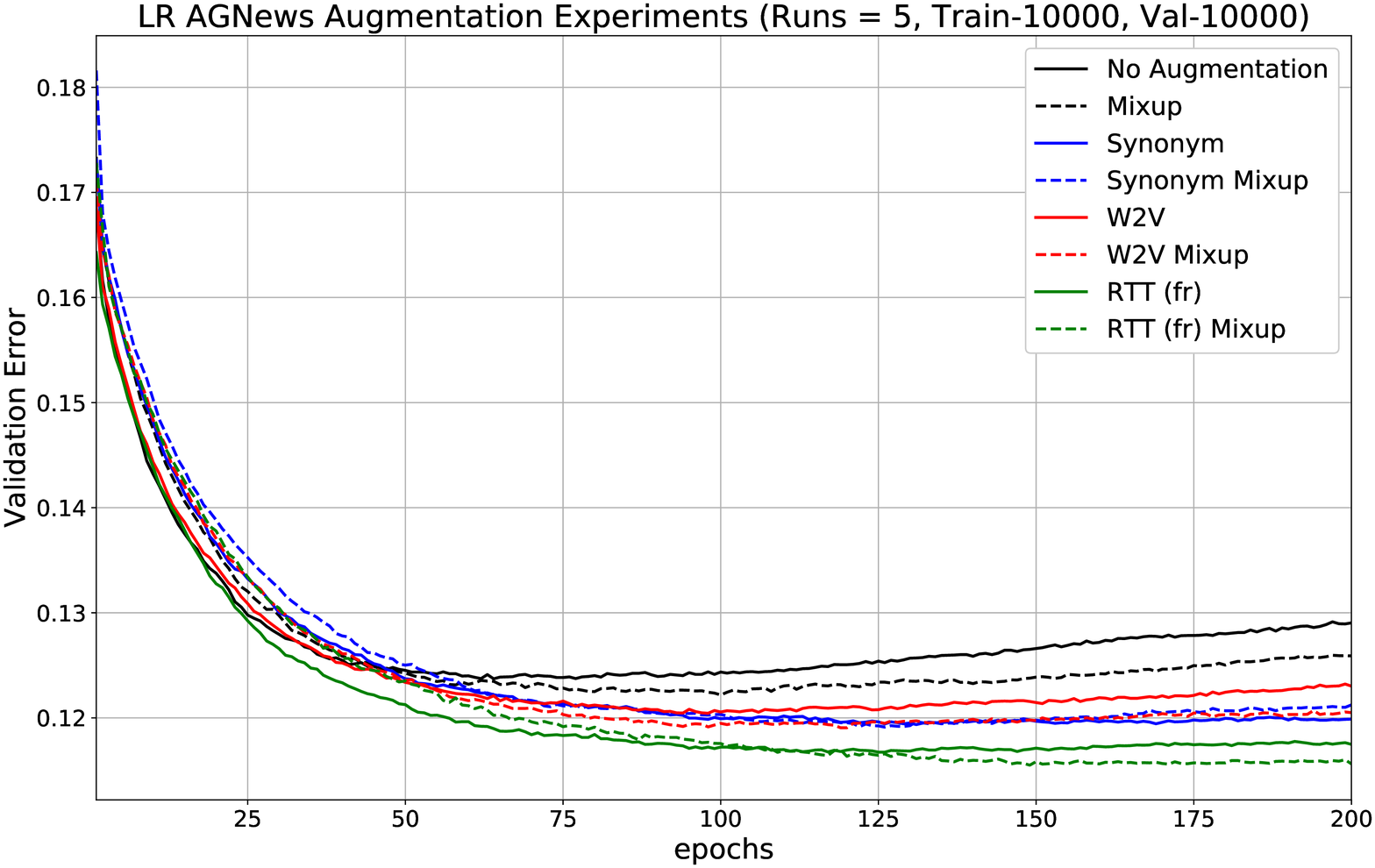}
    }
    \subfloat[AG News - DNN Error \label{fig:augmentation_results_dnn_error_ag_news}]{%
        \includegraphics[width=0.45\textwidth]{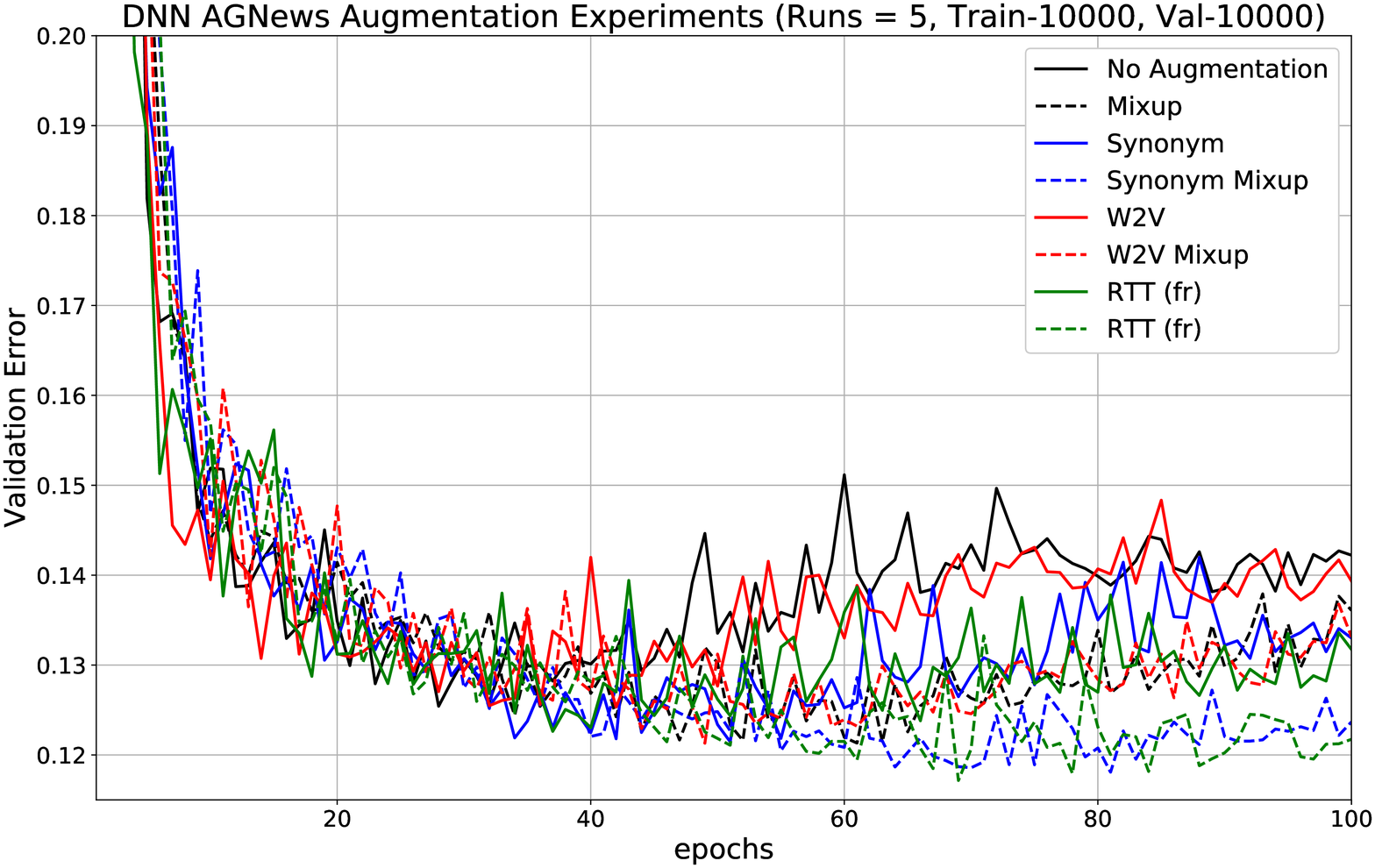}
    }
    \caption{Effect of augmentation on different training set sizes for AG News\label{fig:augmentation_results_ag_news}}
\end{figure}

First looking at the AG News results (Figures \ref{fig:augmentation_results_lr_error_ag_news} and \ref{fig:augmentation_results_dnn_error_ag_news}), we start noticing a few trends. We use 10000 articles for training and a further 10000 for validation. We repeat the experiment 5 times. When looking at the validation error for logistic regression on the AG News dataset, Fig.~\ref{fig:augmentation_results_lr_error_ag_news}, we note that augmented data leads to better results. For the logistic regression results (Fig \ref{fig:augmentation_results_lr_error_ag_news}), the WordNet-based Synonym and W2V augmentation lead to lower validation error. At the same time, across all augmentation schemes, \emph{mixup} improved performance. With the DNN, WordNet-based Synonym augmentation with \emph{mixup} leads to the lowest validation error. We believe this is due to the fact that the AG News dataset is made up of news articles that have longer length and are written formally, as such a synonym database is likely to make good substitutions.

\begin{figure}[ht!]
\centering

    \subfloat[Sentiment 140 - LR error\label{fig:augmentation_results_lr_error_sent140}]{%
        \includegraphics[width=0.45\textwidth]{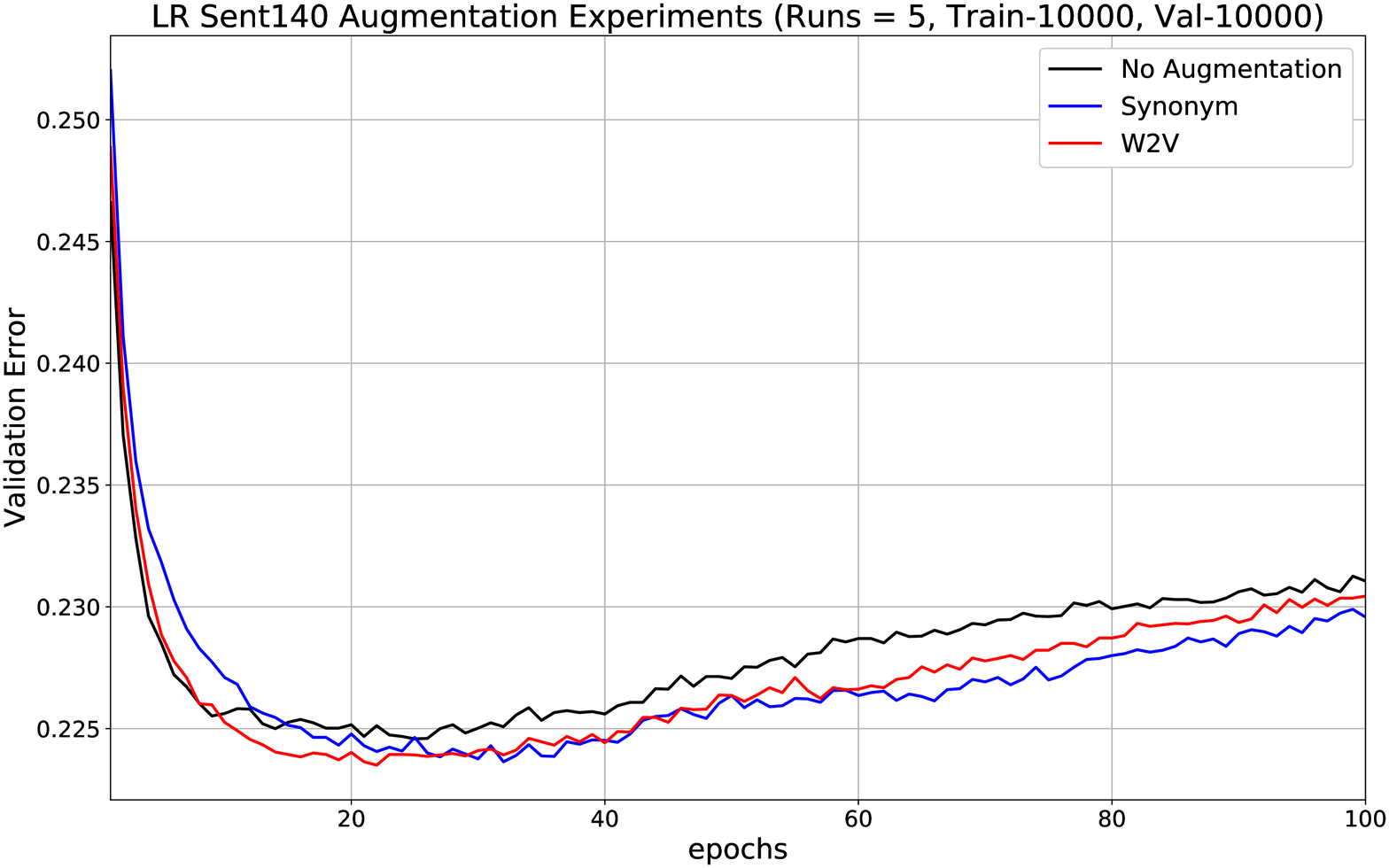}
    }
    \subfloat[Sentiment 140 - DNN Error \label{fig:augmentation_results_dnn_error_sent140}]{%
        \includegraphics[width=0.45\textwidth]{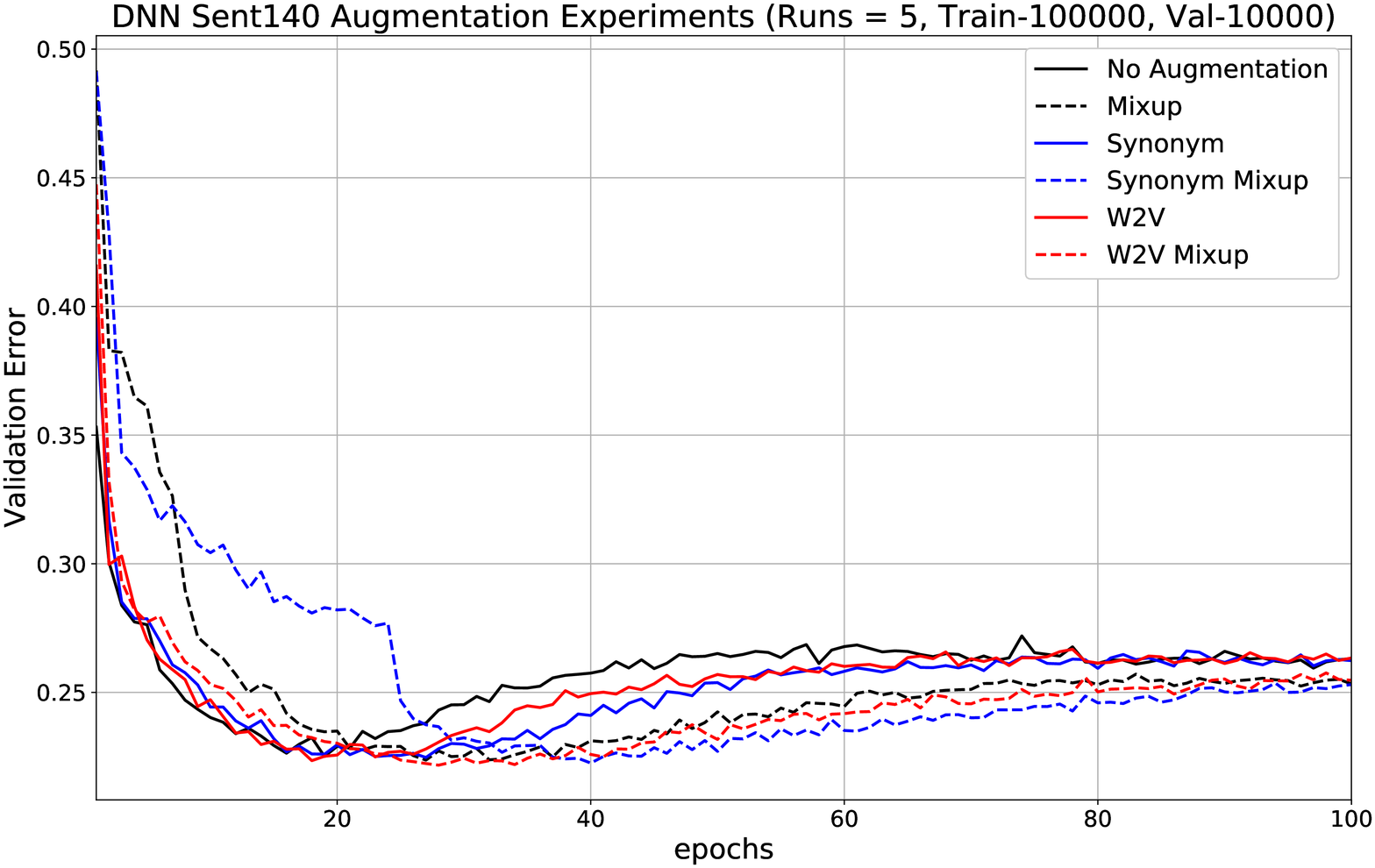}
    }
    \caption{Effect of augmentation on different training set sizes for Sentiment 140\label{fig:augmentation_results_sent140}}
\end{figure}

Turning our attention to the two social media based datasets, we now have to deal with more informal language. In the Sentiment 140 dataset experiments (Figures \ref{fig:augmentation_results_lr_error_sent140} and \ref{fig:augmentation_results_dnn_error_sent140}), we use 100000 posts for training and 10000 for validation. Again augmentation outperforms no augmentation. We are not able to get results for \emph{mixup} on Logistic regression with TFIDF due to computational constraints.  \emph{mixup} still provides better performance, when combined with another augmentation method for the DNN, and we expect that this is the same for LR. \emph{mixup} also affects the performance of the learning even when it is starting to overfit. We discuss more details of this later.

\begin{figure}[ht!]
\centering

    \subfloat[Hate Speech - LR error\label{fig:augmentation_results_lr_error_hatespeech}]{%
        \includegraphics[width=0.45\textwidth]{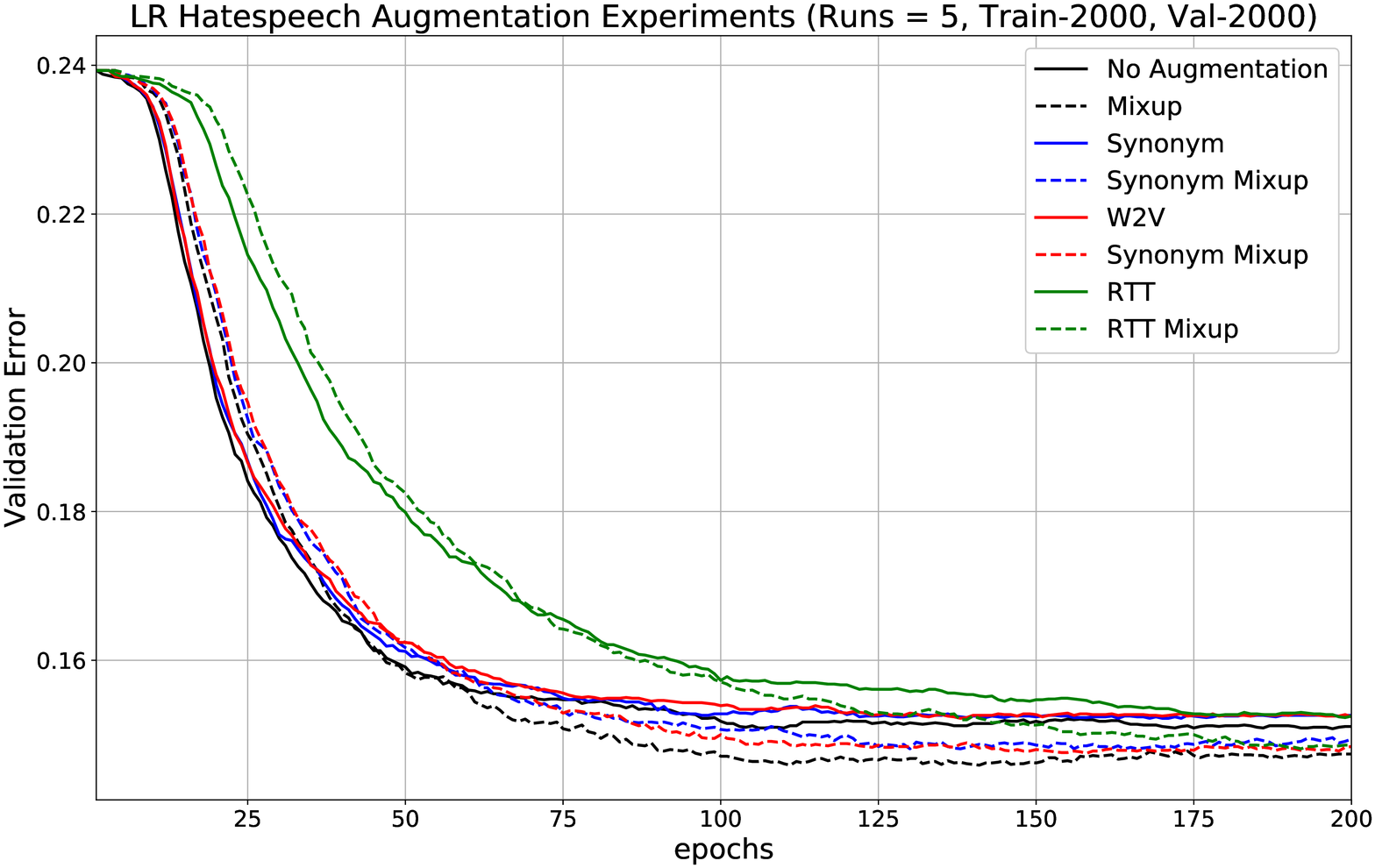}
    }
    \subfloat[Hate Speech - DNN Error \label{fig:augmentation_results_dnn_error_hatespeech}]{%
        \includegraphics[width=0.45\textwidth]{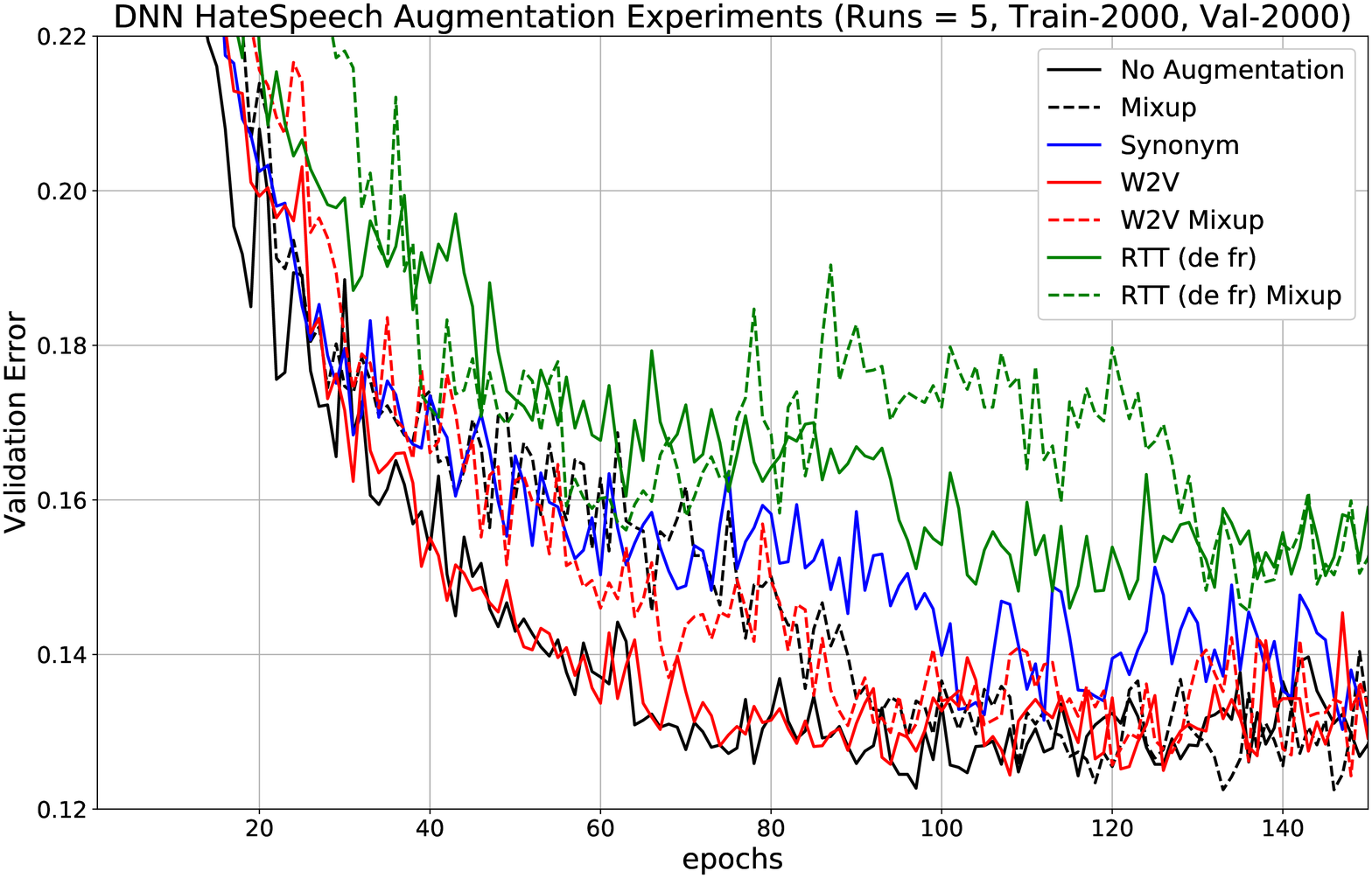}
    }
    \caption{Effect of augmentation on different training set sizes for Hate Speech (Synonym Mixup ommited for readability)\label{fig:augmentation_results_hateSpeech}}
\end{figure}

Finally for the Hate Speech dataset, we use the smallest amount of training data. We have a training set of size 2000 and a validation set of the same size. For the logistic regression (Fig \ref{fig:augmentation_results_lr_error_hatespeech}), the \emph{mixup} augmentation provides good performance. The same result is seen for the DNN (Fig \ref{fig:augmentation_results_dnn_error_hatespeech}). 

RTT Augmentation proved to have differing effects. On the logistic regression, its performance on AG News was the best while on the Hate Speech dataset it was less successful. On the DNN, we observe similar results for both AG News and the Hate Speech data sets. We do think that this does likely indicate that on social media data, RTT will have a hard time doing translations and as such will increase error. At the same time, the more formal AG News dataset benefits from the augmentation. More still needs to be done to explore this area. In the results we present for RTT, for AG News we used Amazon Translate with \emph{English to French and back to English} as well as \emph{English to German and back to English}. For the Hate Speech dataset we used Google translate for \emph{English to French and back to English}.

% we observe that at the low number of epochs the normal data has a lower error. But, with a higher number of epochs (more training time), both Wordnet learning curves are presented for the AG News and Sentiment 140 datasets under different settings. First, we show in Fig.~\ref{fig:augmentation_results_ag_news} what the effects of the augmentation on text are in increasing the data. zzz Discuss result. In the second experiment, we show in zzz ref Fig, the effects of augmentation on representation. Discuss results. Lastly we present in Table~\ref{table:data_size_error} test performance of different combinations, for this we also include results on the Reddit and IMDB datasets zzz. Discuss results.
% \begin{table*}[ht!]
% \centering
% \caption{Test Error Experiments on data size - Logistic Regression \label{table:data_size_error}}
%  \begin{tabular}{|l|l|l|l|l|} 
%  \hline
%  Augmentation Method & AG News & Sentiment 140 & Reddit Sarcasm & IMDB \\ 
%  \hline
%  Original Data [5000] & ? & $0.42\pm0.01$ & ? & ? \\ 
%  Wordnet 5X Augmentation [5000] & ? & $0.41\pm0.01$ & ? & ? \\ 
%  Word2Vec 5X Augmentation [5000] & ? & $0.41\pm0.01$ & ? & ? \\ 
%  RTT [5000] & ? & ? & ? & ? \\ 
%  \hline
%  \end{tabular}
% \end{table*}

\subsection{Effect of different number of augmentations}
Multiple augmentations may have an impact on the error. The next subsections discuss how multiple augmentations impact the error. 
% \subsubsection{WordNet and Word2vec-based augmentation}
We focus on using LR to conduct the experiments. We augment the data in portions of different sizes. For AG News, we augment 1000 and 10000 tokens for 5 and 10 times and report on the error using WordNet-based augmentation. 
% Figure~\ref{fig:num_aug} shows the error against the epochs. 
We obtain the error of 0.1824 when augmenting 1000 tokens for 5 times. The error reduced to 0.1809 when we increase the number of augmentations to 10. This shows the effect of increasing number of augmentation on error is very low. As such, we increase the data size and augment 10k of the data. We observe an error of 0.12136 when augmenting 5 times. Then, we increase the number of augmentations to 10 and the error slightly reduce to 0.121.  With these results for longer text, we observe that number of augmentations has an impact on the error and when augmenting larger data the error is slightly reduced by a difference of 0.12136-0.1824. \par
For Sentiment 140, we augment 10000 tokens using the same settings. We observe an error of 0.25434 when augmenting for 5 times, we increased the number of augmentations to 10 then the error increased to 0.25764, this shows Word2vec introduced more noise on the data. \par

What we observe is that moving from 5 to 10 augmentations had only slight effects on lowest error for the cases we studied.
% with more number of augmentations. 
\begin{figure}[ht!]
\centering
    \subfloat[AG News - LR error \label{fig:lr_num_of_aug_size_error_ag}]{%
        \includegraphics[width=0.4\textwidth]{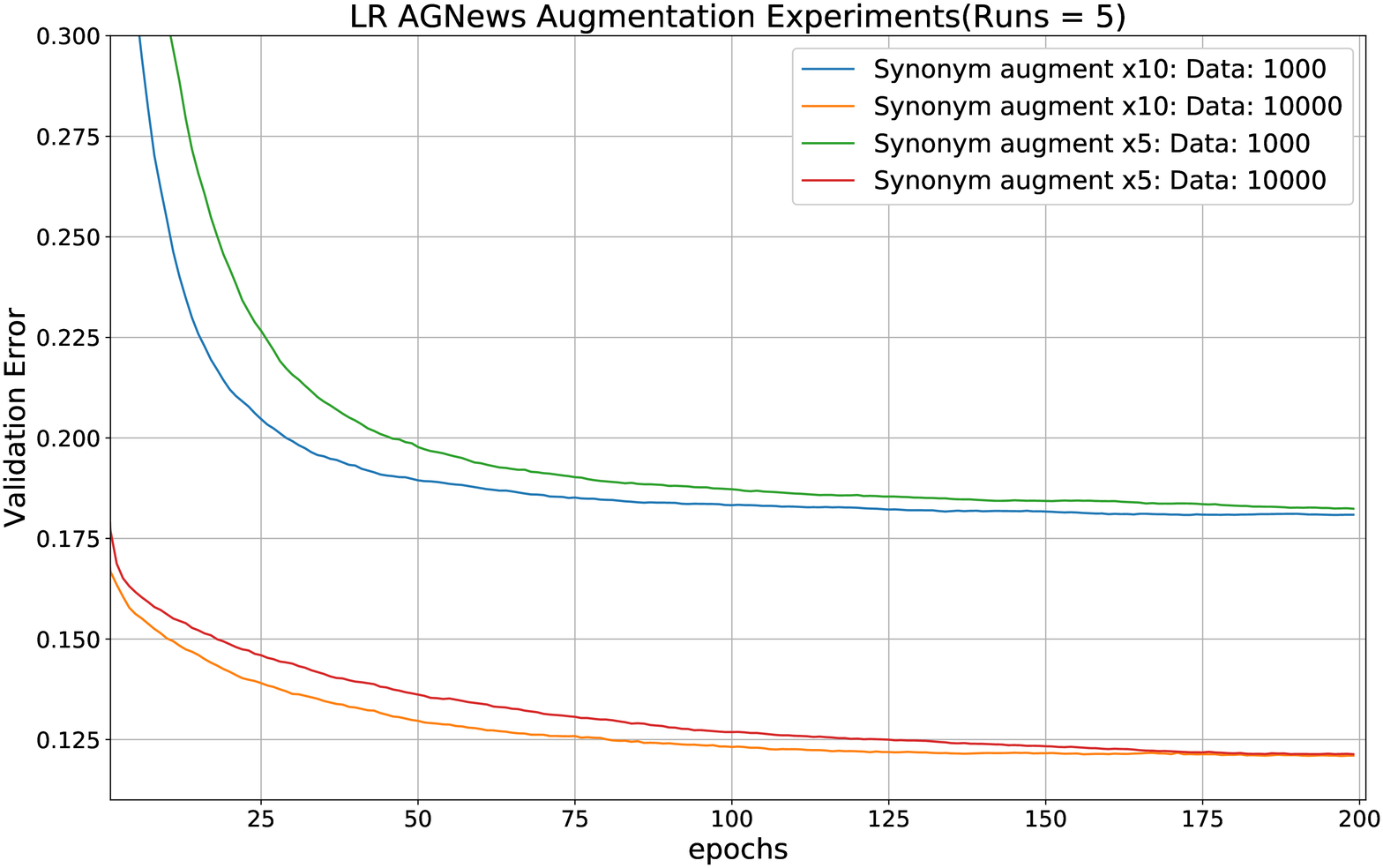}
    }
    \subfloat[Sentiment 140 - LR error \label{fig:lr_num_of_aug_size_error_senti}]{%
        \includegraphics[width=0.4\textwidth]{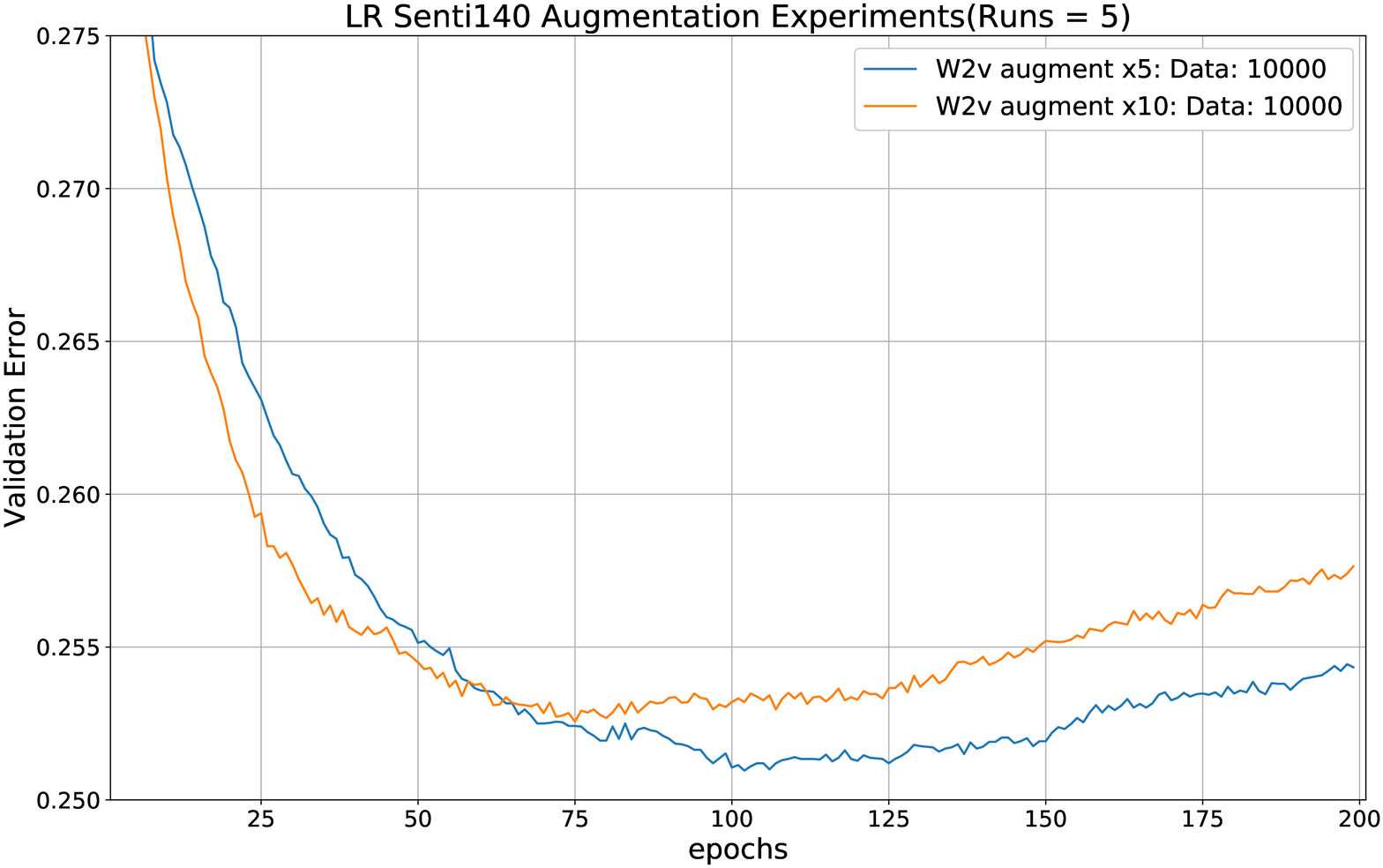}
    }
    \caption{Effects of different sizes of augmentation given different input data on error. (a) WordNet-based augmentation slightly reduces the error with more number of augmentations. (b) Word2vec increase the error by introducing more noise.}
    \label{fig:num_aug}
\end{figure}
% \subsubsection{mixup-based augmentation}
% The above settings are combined with mixup and we observe mixup reducing the error. 
% \begin{figure}[ht]
%     \centering
%     \includegraphics[width=\columnwidth]{mixup_logistic_sizes_error_test.eps}    \caption{\textit{mixup} augmentation on logistic regression with Glove embeddings. Effects of different sizes of augmentation given different input data on error. \textit{mixup} provides lower error with more augmentation }
%     \label{fig:mixup_error}
% \end{figure}

% \begin{figure}[ht]
%     \centering
%     \includegraphics[width=\columnwidth]{mixup_logistic_sizes_loss_test.eps}
%     \caption{\textit{mixup} augmentation on logistic regression with Glove embeddings. Effects of different sizes of augmentation given different input data on loss. \textit{mixup} provides lower error with more augmentation }
%     \label{fig:mixup_loss}
% \end{figure}

\subsection{Effect of augmentation on overfitting} 

\begin{figure}[ht!]
\centering

    \subfloat[AG News - DNN Loss\label{fig:dnn_agnews_mixup_epochs_loss_test}]{%
        \includegraphics[width=0.45\textwidth]{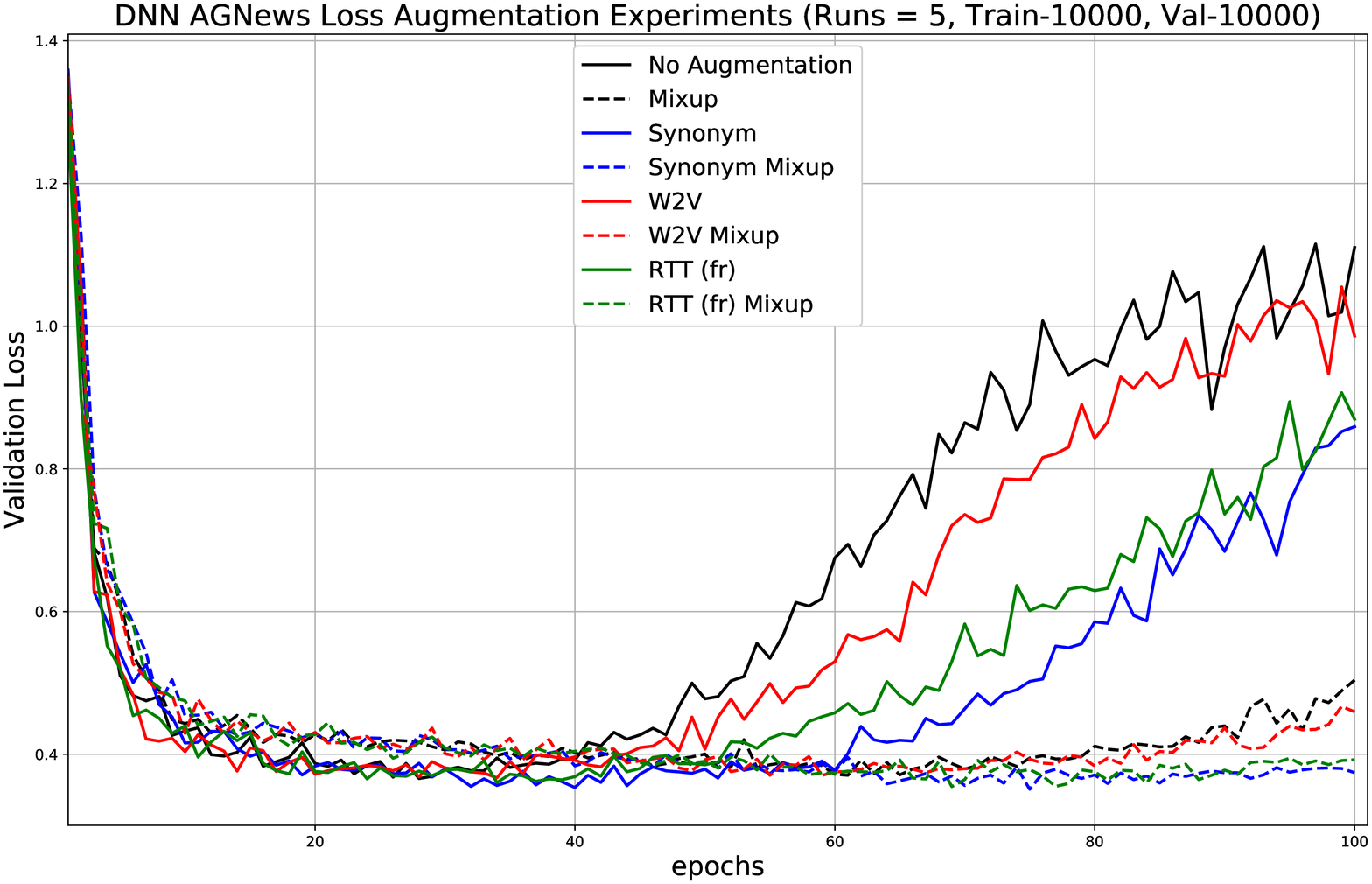}
    }
    \subfloat[Sentiment 140 - DNN Loss \label{fig:dnn_sent140_mixup_epochs_loss_test}]{%
        \includegraphics[width=0.45\textwidth]{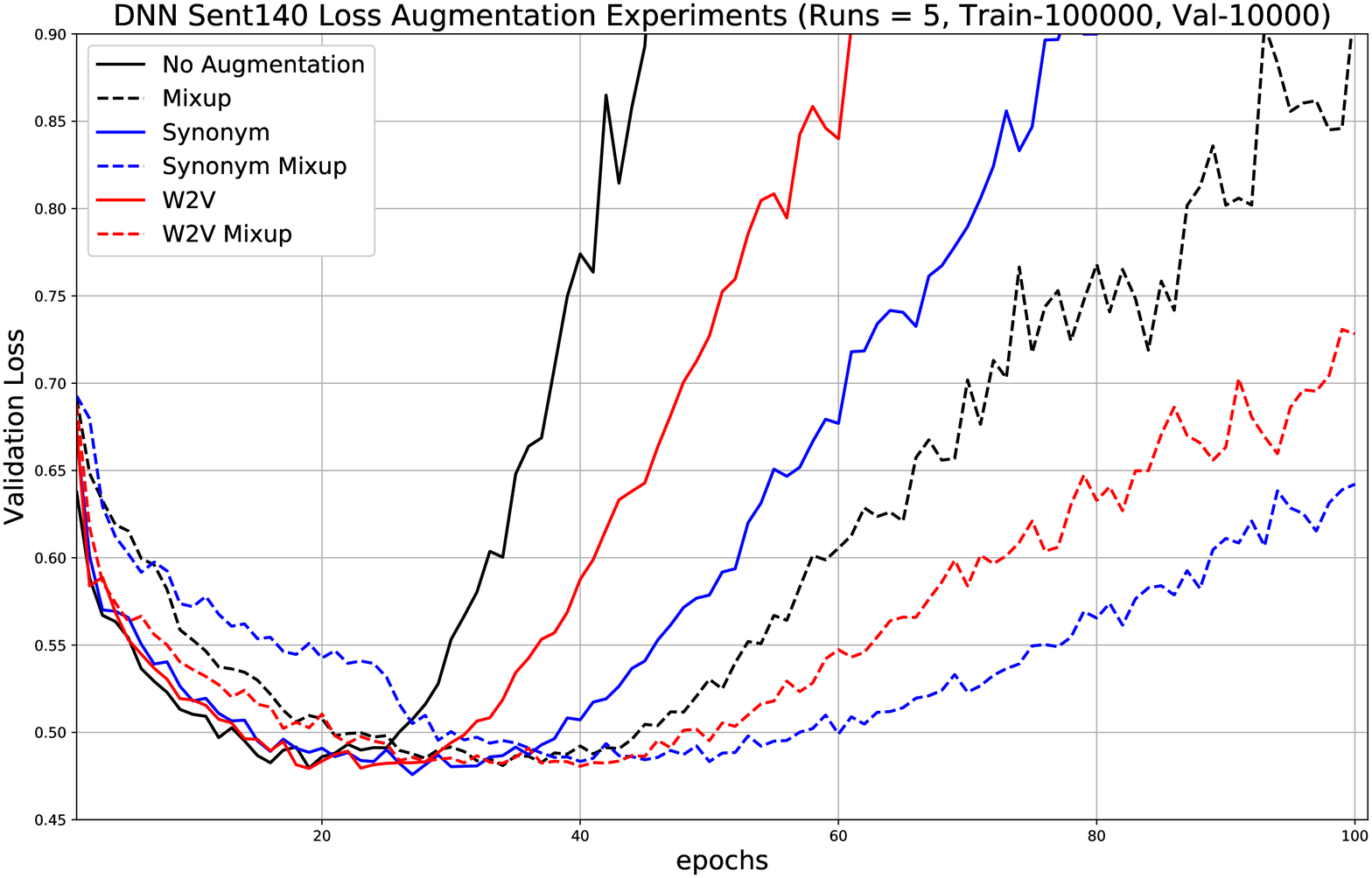}
    }
    \caption{Effect of overfitting on DNN model shown through cross-entropy loss for the AG News and Sentiment 140 datasets \label{fig:loss_plot}}
\end{figure}

Augmentation is viewed as a form of regularisation \cite{smirnov2014comparison}. We can look at the effect of the different augmentation methods on how they reduce the effects of overfitting. Figure \ref{fig:loss_plot} shows the effect of the different methods we tested on overfitting. On both the AG News dataset (Fig \ref{fig:dnn_agnews_mixup_epochs_loss_test}) and the Sentiment 140 dataset (Fig \ref{fig:dnn_sent140_mixup_epochs_loss_test}) augmentation reduces the overfitting. \emph{mixup} has the largest impact. On AG News, overfitting effect is reduced by Synonym, W2V and RTT augmentations, even without \emph{mixup}. 

\section{Conclusion and Future Work\label{sec:conclusion}} 
% The test results error scores on the testing data are shown in Table zzz. 
% As can be seen from the table, 
The use of WordNet-based Synonym augmentation on AG News or long text does result in most of the words found from the database. Hence, the probability of those words to be augmented is higher and the resulting augmented data does not change the meaning of the message. And this, results in WordNet-based augmentation approach being a rich augmentation compared to RTT and Word2vec-based approach. The WordNet-based augmentation approach highly depends on an existing database. Hence, the disadvantage is when such database in not available for low-resourced languages, it first needs to be created at a great cost. The alternative way of augmenting low-resource language, is to use unsupervised word embedding models that can be trained using Glove, Word2vec, and fastText \cite{joulin2017bag} on pre-collected available corpuses such as Wikipedia, Newspapers or literature. The vector representations in the models can be used to identify the nearest neighbours via the cosine similarity (such as used in the the Word2vec based augmentation). Such an approach becomes more feasible for augmenting data from lower resourced languages. As shown in the paper experiments Word2vec-based augmentation resulted with good comparable results compared to synonym-based approach on Sentiment 140, this shows that augmentation can be done with only Word2vec. Even on AG News, Word2vec-based augmentation is competitive.

RTT-based augmentation is expensive in many ways. If using an online service, the commercial services available require financial resources. The free tiers made available will only be able to translate a few thousand words for free. If one wants to reduce the cost, then one can train or use a pre-trained neural machine translation model. Commercial strength grade translation models though are hard to come by and will require a lot of data to train (which is another cost). As such they are even less feasible for lower resourced languages. We were only able to use RTT augmentation on the smaller datasets of AG News and Social Media Hate Speech. Even with this, we had to use two different services (Google and Amazon) to keep costs low. In our academic setting it is not feasible to use RTT on very large datasets and remains future work. 

There are a number of avenues of future work. We have provided experiments showing efficacy of different augmentation schemes on a level playing field. More local context augmentation using language models is an avenue of extending this work. Another avenue is investigating how to improve semi-supervised learning of low resourced languages using augmentation. Given the success of \emph{mixup}, one can explore other methods that augment the data as a way of regularisation.   

\bibliographystyle{splncs04}
\bibliography{references}




% LNCS Homepage, \url{http://www.springer.com/lncs}. Last accessed 4
% \end{thebibliography}
\end{document}